\documentclass[10pt,twocolumn,letterpaper]{article}
\usepackage[algorithms]{wacv} 

\usepackage{amsmath}
\usepackage{graphicx}
\usepackage{amssymb}
\usepackage{booktabs}
\usepackage{comment}
\usepackage{microtype}
\usepackage{graphicx}
\usepackage{booktabs}
\usepackage{subcaption}
\usepackage[export]{adjustbox}
\usepackage{multirow}
\usepackage{xcolor}
\usepackage[accsupp]{axessibility}

\DeclareMathOperator*{\argmax}{arg\,max}
\DeclareMathOperator*{\argmin}{arg\,min}
\newcommand{\model}{CoSeD\xspace}

\newcommand*{\email}[1]{\texttt{#1}}

\usepackage[pagebackref,breaklinks,colorlinks]{hyperref}

\usepackage[capitalize]{cleveref}
\crefname{section}{Sec.}{Secs.}
\Crefname{section}{Section}{Sections}
\Crefname{table}{Table}{Tables}
\crefname{table}{Tab.}{Tabs.}

\def\wacvPaperID{2590} 
\def\confName{WACV}
\def\confYear{2025}

\begin{document}

\title{Contrastive Sequential-Diffusion Learning:\\ Non-linear and Multi-Scene Instructional Video Synthesis}

\author{
 \textbf{Vasco Ramos\textsuperscript{1}},
 \textbf{Yonatan Bitton\textsuperscript{2}},
 \textbf{Michal Yarom\textsuperscript{2}},
 \textbf{Idan Szpektor\textsuperscript{2}},
 \textbf{Joao Magalhaes \textsuperscript{1}}
\\
 \textsuperscript{1}NOVA LINCS, NOVA School of Science and Technology, Portugal
 \\
 \textsuperscript{2}Google Research
\\
\email{jmag@fct.unl.pt},
\email{szpektor@google.com}
}
\maketitle

\begin{abstract}
Generated video scenes for action-centric sequence descriptions, such as recipe instructions and do-it-yourself projects, often include non-linear patterns, where the next video may need to be visually consistent not with the immediately preceding video but with earlier ones.
Current multi-scene video synthesis approaches fail to meet these consistency requirements.
To address this, we propose a contrastive sequential video diffusion method that selects the most suitable previously generated scene to guide and condition the denoising process of the next scene. 
The result is a multi-scene video that is grounded in the scene descriptions and coherent w.r.t. the scenes that require visual consistency.
Experiments with action-centered data from the real world demonstrate the practicality and improved consistency of our model compared to previous work. Code and examples available at \href{https://github.com/novasearch/CoSeD}{https://github.com/novasearch/CoSeD}
\end{abstract}

\section{Introduction}
When people perform tasks involving numerous intricate steps, complementing textual instructions with visual illustrations enhances the user experience~\cite{grifoni2009multimodal,serafini2014reading}. For this reason, various platforms and tools provide multi-scene videos to convey instructional content, such as recipe instructions and do-it-yourself (DIY) projects~\cite{lin-etal-2020-recipe}.

\begin{figure}[t]
    \centering
    \includegraphics[width=\linewidth]{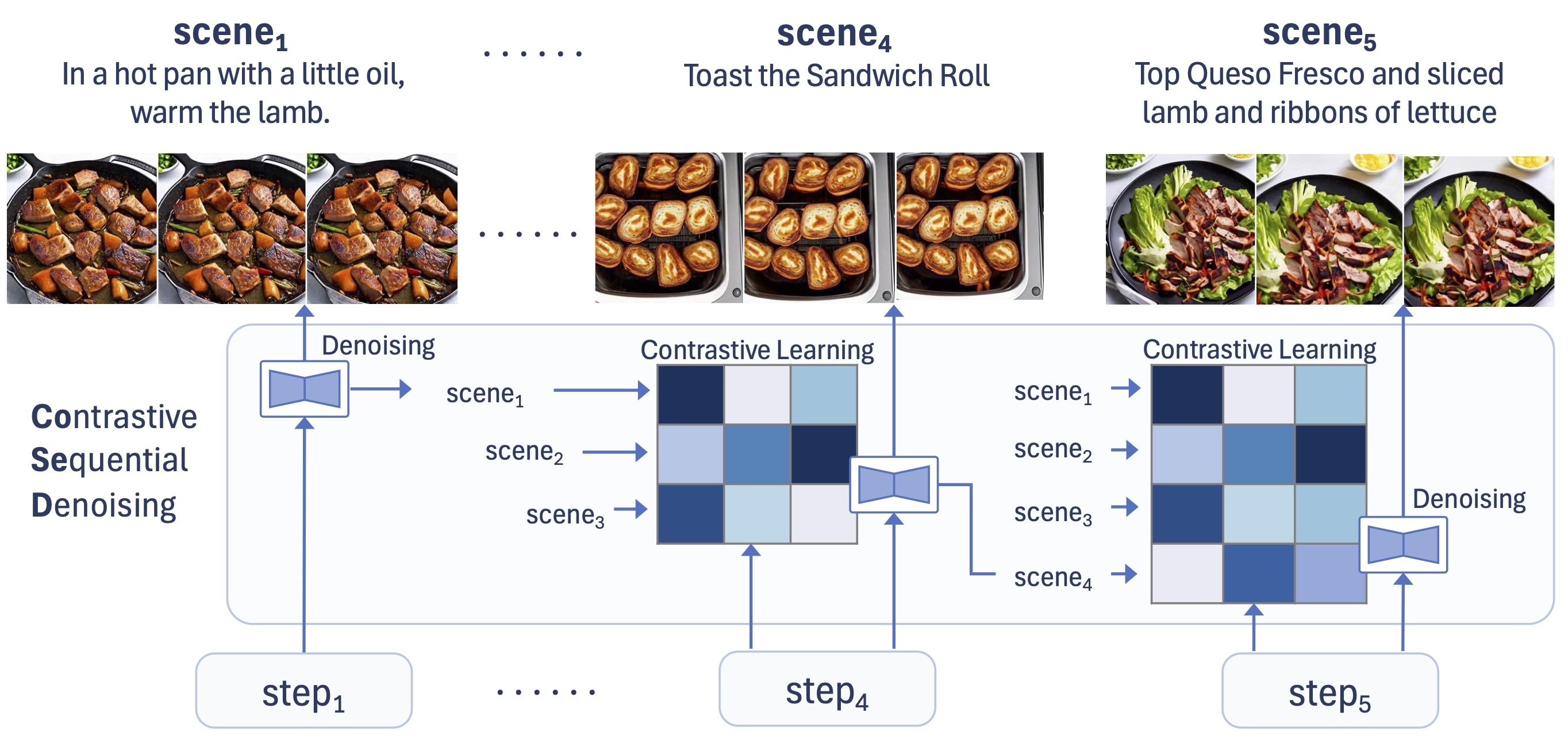}
    \caption{\model is grounded on an input sequence of step actions to synthesize non-linear, multi-scene instructional videos.} 
    \label{fig:scene-generation}
\end{figure}

State-of-the-art video synthesis methods demonstrate remarkable performance in generating single-scene videos~\cite{imagen-video,make-a-video,lavie,stable_video_diffusion,align_your_latents}. Yet, only a few works address multi-scene video generation~\cite{video_drafter,videodirectorgpt,mora}. 
These methods focus on domains where a single character is central to all scenes, achieving coherence by reusing and combining visual elements across frames. However, multi-scene instructional video synthesis raises a number of challenges.
First, the input is a \textit{strict sequence of actions}, for which it is necessary to \textit{generate the full sequence of videos}. A model should not generate just the last scene, like GILL~\cite{gill}, and one cannot provide the topic and let the model generate a random sequence of text-video pairs similarly to VideoDrafter~\cite{video_drafter}.
Second, similar to story generation~\cite{AR-LDM,ACM-VSG,make-a-story,StoryDALL-E}, the generative model needs to \textit{determine which previous step in the sequence to use as the basis for grounding each new scene}.
Third, while existing methods are focused on video generation where a single (typically human) character is the center of all scenes~\cite{video_drafter}, instructional videos typically incorporate multiple objects instead of central characters. 
Hence, we argue that multi-scene instructional video synthesis requires an approach that is sequence-grounded by design, (see Figure~\ref{fig:scene-generation}).

To this end, we propose \model (\textbf{Co}ntrastive \textbf{Se}quential \textbf{D}iffusion learning), a novel approach to instructional video generation. Our method generates candidate images for each step based on the textual description and latent information from previous steps. We then use a contrastive selection approach to choose the best image by evaluating it against prior step descriptions and images. Finally, we use these images to produce a video for each step, ensuring an accurate and consistent representation of the entire task sequence.

\model was able to generate coherent video sequences across diverse instructional content, maintaining high fidelity and relevance in aligning language with vision. 
\model showed a 20\% improvement in human evaluations compared to existing multi-scene methods and was preferred 68\% of the time in side-by-side comparisons.
Additionally, the compact size of our model enables efficient training and fine-tuning. To summarize our contributions:
\begin{itemize}
    \item To the best of our knowledge, we are the first to address multi-scene instructional video generation.
    \item We introduce contrastive diffusion learning over latents sampled from previous generations.
    \item We contribute to a better understanding of the role of seeds and the conditioning of the reverse diffusion process in prior latent representations.
\end{itemize}

\section{Related Work}

Various approaches have been explored to address coherence in image generation. AR-LDM \cite{AR-LDM} introduces a history-aware autoregressive latent diffusion model that incorporates information from previous steps into the diffusion model's cross-attention mechanism to guide generation. However, achieving the reported results requires intensive training of the entire pipeline for each dataset, which includes 650 million parameters. Make-a-Story~\cite{make-a-story} incorporates the complete history of intermediate image representations (latent vectors), which may introduce noise and potentially lead to content generation based on less relevant past information. GILL~\cite{gill} fuses frozen text-only large language models (LLMs) with pre-trained image encoder and decoder models through a mapping network, enabling multimodal capabilities like image retrieval and generation. However, this can break coherence if retrieved images do not align with the context. SEED-LLaMA~\cite{seed-llama} integrates a visual tokenizer with a multimodal LLM to process text and images, excelling in multi-turn generation, but struggles with maintaining narrative coherence in story generation tasks.

To generate long single-scene videos Blattmann~\cite{align_your_latents} and Yin~\cite{nuwa-xl} generate sparse key frames and interpolate intermediary frames recursively to enhance the frame rate. Extending this idea, Stable Video Diffusion~\cite{stable_video_diffusion} creates a large dataset of annotated video clips by filtering out those with low motion or excessive text, resulting in the generation of higher quality videos. In contrast, Lumiere~\cite{lumiere} generates the entire video in a single pass, eliminating the need for sparse key frames and interpolation.

Improvements have also been made in generating coherent multi-scene videos. Video Drafter~\cite{video_drafter} employs brute-force LLM prompting to create distinct scenes and detailed descriptions for each element. Then it generates image templates that are combined with scene descriptions to produce the final video. Similarly, VideoDirectorGPT ~\cite{videodirectorgpt} employs a two-stage process in which GPT-4~\cite{gpt4} expands text prompts into detailed descriptions and ensures visual continuity by generating textual descriptions and entity layouts. Mora~\cite{mora} introduces a multiagent framework, breaking tasks into subtasks like refining prompts~\cite{llama}, and generating images to create small video segments, which are then assembled~\cite{seine} into a coherent final video, achieving performance comparable to closed-source models such as SORA~\cite{sora}. However, it relies on using only the last frame of each video segment to start the generation of the next one, which can be problematic if the current step is not directly related to the previous one. StoryDiffusion~\cite{storydiffusion} maintains coherence across frames using a self-attention mechanism and a module for smooth transitions, ensuring that videos faithfully depict the input prompt. Lastly, TALC~\cite{talc} enhances text-to-video (T2V) models by improving the temporal alignment between video scenes and text segments, improving visual fidelity and narrative coherence.

Building on these advancements, we develop a method that addresses the challenge of maintaining scene coherence while respecting text descriptions, while keeping the model compact to facilitate fine-tuning across multiple domains.

\begin{figure*}[t]
    \includegraphics[width=\linewidth]{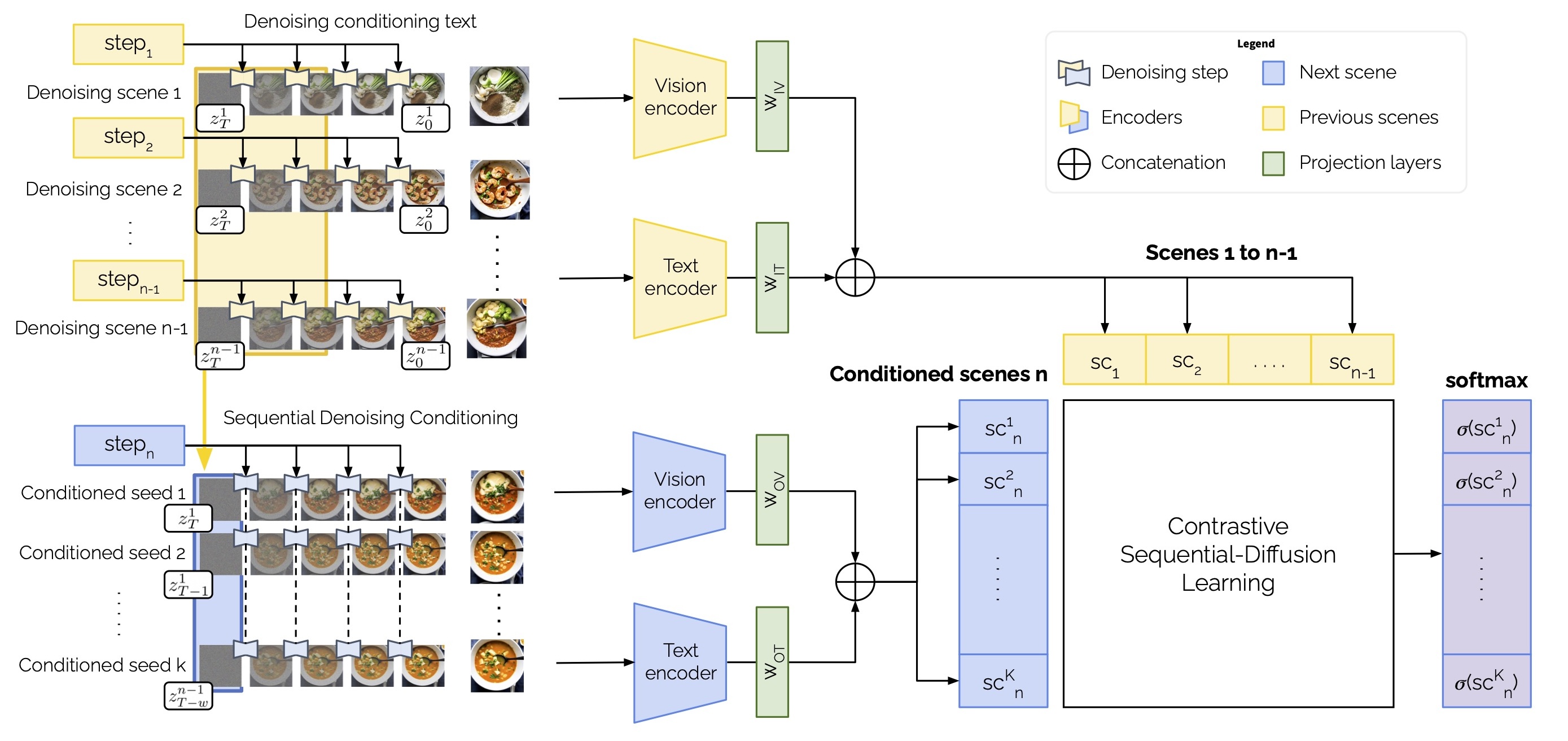}
    \caption{The proposed contrastive denoising diffusion learning architecture. The contrastive learning component captures the temporal relationships between conditioned scenes and preceding scenes, ensuring coherent transitions throughout the video.}
    \label{fig:architecture}
    \vspace{-3mm}
\end{figure*}

\section{Problem Setting}
This section outlines our methodology for generating scene sequences that align with each instruction while preserving continuity with preceding scenes. By enhancing Latent Diffusion Models, we can effectively learn the interdependencies across scenes and guarantee a cohesive visual progression, even when step relations are non-linear.

\subsection{Sequential Scene Dependency}
Given a set of tasks, $\mathcal{D} = \{ T_1, T_2, ...\}$, where each task $T_j$ comprises a sequence of step-by-step text instructions $T_j=\{s_{j_1}, \ldots, s_{j_n} \}$, our goal is to generate the sequence of scenes $V_j=\{v_{j_1}, \ldots, v_{j_n}\}$ that are best aligned with the corresponding step instruction and all previous visual scenes. The result is a multi-scene video that depicts the steps of the task consistently across all scenes. For simplicity, we will omit the task index $j$ from our notation.

In our setting, we depart from the linear dependency assumption used in previous works~\cite{mora} and acknowledge the possibility of a more complex and non-linear sequential structure~\cite{donatelli-etal-2021-aligning}. To address this assumption, the model needs to consider not only the current step description and visual scene pair ($s_n$, $v_n$), but also the pairs from the previous steps and visual scenes, $\{(s_1,v_1), \ldots, (s_{n-1},v_{n-1})\}$. This ensures coherence in the visual elements generated, maintaining consistency, and reflecting the progression of the task, even when individual steps are vague or missing details.

\subsection{Sequential Multi-Scene Diffusion}
Latent Diffusion Models are designed to synthesize one single image or video at a time. Our goal is to move beyond this limitation and propose a sequential-diffusion model that learns how semantic and visual dependencies should exist in a sequence of multiple scenes.
Using the Latent Diffusion Models formulation proposed by~\cite{stable-diffusion}, the independent denoising process for each isolated step $s_n$ of a sequence is the direct application of the model,
\begin{equation}
    \mathcal{L}_{LDM} = \mathbb{E}_{z_t^n, s_n,\epsilon, t} \Bigl[ \| \epsilon - \epsilon_\theta(z_t^n, t, \tau_\theta(s_n)) \| ^2_2 \Bigr]
\end{equation}
where $z_t^n$ corresponds to the denoising iteration $t$ of the visual scene $v_n$, hence $z_0^n = E(v_n)$.
Formally, we wish to learn a sequence model that iteratively estimates the $v_n$ scene that maximizes the likelihood given the entire sequence of all previous $n-1$ steps. Formally, we have,
\begin{equation}
    \sum_{n=2}^N p(v_{n}|s_{n}, (s_{n-1},v_{n-1}),\ldots(s_1,v_1)),
\end{equation}
where $N$ is the total number of steps in a given sequence.
We propose to ensure sequential consistency through the text conditioning encoder $\tau_\theta(s_i)$ and the visual denoising seed $z^n_T$.
The proposed contrastive sequential diffusion learning, Figure~\ref{fig:architecture}, incorporates these two methods and will be discussed in the next section.

\section{\model: Contrastive Sequential Diffusion}
Our approach aims to find the most accurate image for generating a video that depicts a step of a task. We begin by using a text decoder model to better align the step description with a visual caption/prompt (Section~\ref{sub:language}). Then we generate candidate images based on the information from previous images and the visual caption of the current step (Section~\ref{sub:latents}). Next, we use a contrastive selection method to choose the most suitable candidate image for video generation (Section~\ref{sub:contrastive-selection}). This involves encoding both the step description and the visual scene (Section~\ref{sub:embeddings}). Finally, we use the sequential information of the tasks to train our model to accurately select the most coherent image (Section~\ref{sub:training}).

\subsection{Sequential Language Conditioning}
\label{sub:language}
Following the work of Bordalo et al.~\cite{bordalo24}, we use an LLM to transform the sequence of text descriptions of each step into visual captions. This has been shown to produce text-to-image prompts that are visually richer, leading to better results~\cite{bordalo24, ella}.
Hence, we train a decoder model $\varphi$ to convert the entire context into one self-contained description, 
\begin{equation}
    \varphi(s_n | \{ s_{n-1}, \ldots s_{1}\}),
\end{equation}
whose output is used to condition the denoising process on the entire sequence of actions, leading to the loss function
\begin{equation}
    \mathcal{L}_{CoSeD} = \mathbb{E}_{z_t^n, s_n,\epsilon, t} \Bigl[ \| \epsilon - \epsilon_\theta(z_t^n,t, c_n) \| ^2_2 \Bigr],
    \label{eq:loss_sld}
\end{equation}
where $c_n=\tau_\theta(\varphi(s_i|s_{<i}))$ is the conditioning embedding vector that is passed to the cross-attention of the U-Net $\epsilon_\theta$. 

\subsection{Sequential Denoising Conditioning}
\label{sub:latents}
Previously, we discussed the sequential dependency assumption $(s_n | s_{n-1}, \ldots s_{1})$ for the input of the denoising process. However, achieving sequential dependency within the denoising process itself poses a challenge. Although aligning the input description with the desired output enhances the final result, it does not inherently enforce the generation of visually consistent images. This can lead to accurate depictions of steps but without visual coherence. Therefore, guiding the denoising process is essential to ensure visual coherence across sequential outputs.

We propose a contrastive method to select the image that best represents $s_n$ in terms of its description and the preceding scenes. This selective approach strikes a balance between conditioning in all pairs~\cite{AR-LDM}, which provides comprehensive information but can be slow and difficult to train, and conditioning on only a single preceding step~\cite{bordalo24}, which offers a faster but less detailed approximation. Our method effectively captures the non-linear nature of the steps in the task that we aim to model.
Formally, the denoising iterations follow equation~\ref{eq:loss_sld}, except for the starting iteration $T$ of the reverse diffusion process, 
\begin{equation}
    \mathcal{L}_{CoSeD} = \mathbb{E}_{z_T^n, s_n,\epsilon, t=T} \Bigl[ \| \epsilon - \epsilon_\theta(z_T^n, c_n) \| ^2_2 \Bigr],
    \label{eq:loss_csd}
\end{equation}
where instead of initializing the latent variable $z_T^n$ with a random sample from $z_T \sim \mathcal{N}(\mu,\,\sigma^{2})$, we propose to sample denoised latents from prior steps $s_{<n}$ in the sequence. Formally, for each step $n$, we consider the set of latents produced in previous denoising iterations of earlier steps, i.e.
\begin{equation}
    \{z_T^{i}, z_{T-1}^{i}, \ldots, z_{T-w}^{i} \}_{}^{i\in(1, \ldots, n-1)},
\end{equation}
where $i$ indexes all steps from $1$ to $(n-1)$ and $w$ is the window size over the first denoising latents of each step.
Finally, all candidate visual scene $v_n^i$ are generated with the set of latents. By conditioning the generation of step $n$ on latents from all previous steps in this complex non-linear way, the coherence of the generated sequence is improved.

This method allows for the selection of the most suitable latent representations for each step, ensuring coherence and continuity throughout the entire generation process.

\subsection{Multi-Scene Contrastive Selection}
\label{sub:contrastive-selection}
\paragraph{Text and Vision Scene Embeddings.}
\label{sub:embeddings}
To effectively handle the text and visual modalities of a scene $sc_n=(s_n, v_n)$ in a sequence, we encode both modalities using CLIP~\cite{clip}, see Figure~\ref{fig:architecture}. 
Subsequently, the output of each encoder, is linearly projected to reduce its dimension to half the original size.
For the projection, we use four distinct weight matrices: $W_{IT}$ and $W_{OT}$ for the text embedding, and $W_{IV}$ and $W_{OV}$ for the visual embedding. $W_{IT}$ and $W_{IV}$ project the embeddings of past scenes, while $W_{OT}$ and $W_{OV}$ project the embeddings of the current scene. The projected embeddings are then concatenated into one single vector.

The resulting embedding projections $sc_n^i$ of all candidate scenes $(s_{n},v_{n}^i)$ and all past scenes $sc_{<n}$ allow us to represent all scenes within a unified embedding space.

\begin{figure}[t]
    \centering
    \includegraphics[width=\linewidth]{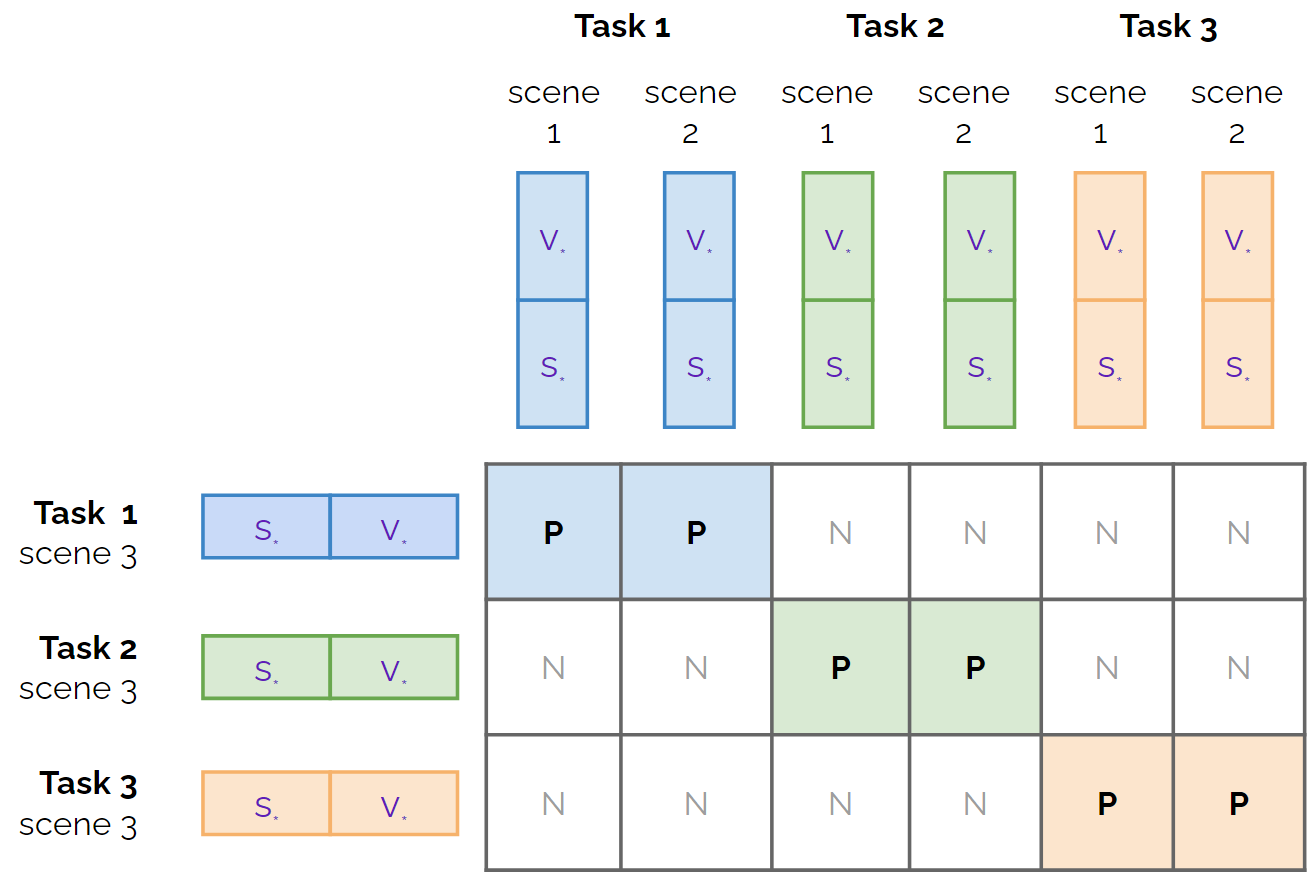}
    \caption{Multi-scene V\&L contrastive learning uses multiple sequences. This multi sequence information serves as both positive and negative pairs helping the model to learn the best next scene according to the ground-truth scenes.}
    \label{fig:training}

\end{figure}

\paragraph{Contrastive Selection.}
We prioritize the visual representation that achieves the best overall consistency throughout the sequence. This selection is achieved by comparing the conditioned scenes with the previous scenes, ensuring that the final output maintains visual coherence.
To achieve this, we first represent a scene $sc_n = (s_n,v_n)$ as the concatenation of its text and visual embeddings, and then calculate the dot product between each conditioned scene and the previous scenes, Figure~\ref{fig:training}. This allows us to measure the similarity between the conditioned scene in step \( n \), denoted as \( sc_n \), and all preceding scenes \( sc_{<n} \). We define this similarity as
\begin{equation}
    \sum_{k=1}^{n-1} sc_{n} \cdot sc_{k}.
\end{equation}

Next, we apply the softmax function to these similarity scores to convert them into probabilities, making it easier to compare how each conditioned scene is related to the previous scenes.
Finally, we select the conditioned scene \( sc_n^i \) with the highest probability of generating the video for the next step. This corresponds to computing the
\begin{equation}
\argmax_{v_{n}^i} \sigma_{\text{\model}}(sc_n^i, sc_{<n}),
\end{equation}
where \( \sigma_{\text{\model}} \) is the softmax function applied over all conditioned scenes, and \( v_n^i \) is the image associated with the selected conditioned scene $sc^i_n = (s_n,v^i_n)$.

By following this process, we ensure that each step in the video is generated based on the conditioned scene that has the strongest visual and contextual relationship to the previous steps, optimizing the flow of the overall sequence.

\paragraph{Contrastive Training.}
\label{sub:training}

During the contrastive selection training phase, we fine-tune a 600K-parameter model to learn the relationships between sequential steps across multiple tasks simultaneously. The model processes a set of $N$ steps (both descriptions and images) from a pool of $M$ tasks. For each task, the model is given a step to be processed in the 'next scenes', while all preceding steps of that task, located in the 'past scenes', serve as context. This setup allows the model to effectively leverage sequential dependencies and learn how each future step relates to its corresponding past steps within the same task, thus improving coherence (see Figure~\ref{fig:training}). Formally, we adopted the cross-entropy loss function,
\begin{equation}
    \argmin_{w_*} \sum_{t}^M \sum_{k=1}^{N} l_{t,k} \log \sigma_{CoSeD}(v{_{n}^j} , s_{n})
    \label{eq:p_seq}
\end{equation}
to guide the learning process, by comparing the model's predictions $\sigma_{CoSeD}(\cdot)$ with one-hot encoded ground truth labels $l_{t,k}$ for each task $t$ and step $k$. These ground truth labels indicate whether a specific step belonged to a task represented in the context. More details can be found in the appendix file.

\section{Experimental Setting}
In this section, we describe the experimental setup used to evaluate the performance of \model in generating multi-scene video and image sequences for manual tasks. We provide details on the dataset used, the backbone models, and the baselines chosen for comparison. The aim is to demonstrate \model's ability to generalize across different models and generate coherent task-oriented outputs.

\paragraph{Dataset.}
\label{sub:dataset}

We used a dataset~\cite{bordalo24} consisting of publicly available manual tasks in recipes and DIY domains.
Each manual task has a title, a description, a list of ingredients, resources, and tools, and a sequence of step-by-step instructions, which may or may not be illustrated. Details about the dataset can be found in the appendix file.

\paragraph{Video Diffusion Backbone Models.}
Since \model is independent of the video generation method, we experimented with both Stable Video Diffusion~\cite{stable_video_diffusion} and Lumiere~\cite{lumiere} models for multi-scene video generation. Stable Video Diffusion was selected for its public availability, while Lumiere was chosen for its enhanced capability to represent complex motion effectively.

\paragraph{Baselines.}
To evaluate the effectiveness of \model in generating coherent image and video sequences for real-world manual tasks, we compared its performance against existing approaches: TALC~\cite{talc} with ModelScope~\cite{modelscope} and with Lumiere~\cite{lumiere}, SD 2.1~\cite{stable-diffusion} with Stable Video Diffusion~\cite{stable_video_diffusion}, stand-alone Lumiere~\cite{lumiere}, and for image sequences we tested Gill~\cite{gill} and Seed-LLama~\cite{seed-llama}.
During the evaluation, we prompted all models to generate a complete task.

\begin{figure*}[t]
    \centering
    \begin{minipage}{0.47\linewidth}
        \centering
        \includegraphics[width=0.9\linewidth]{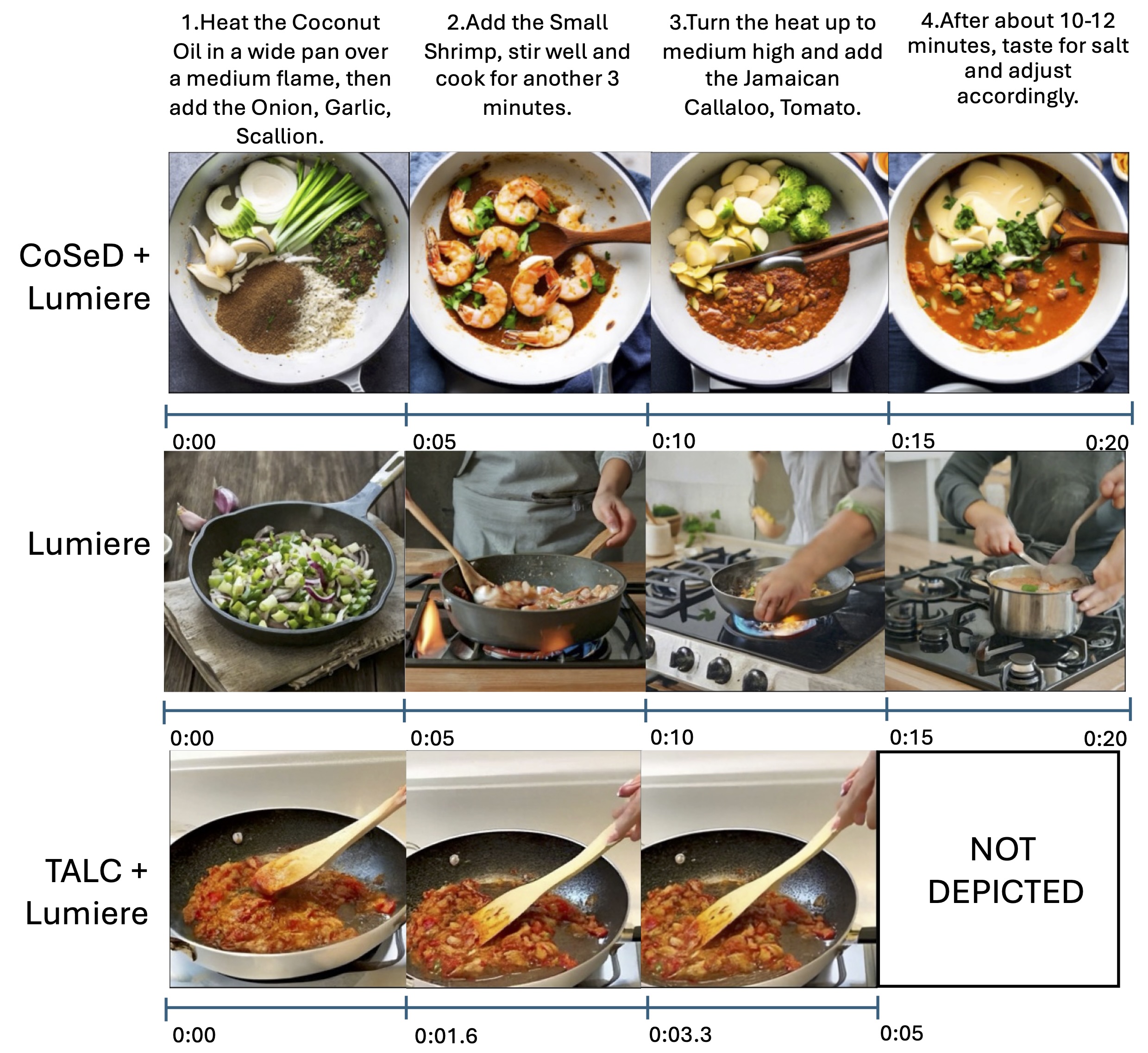}
        \caption{Example of an illustration for the recipe domain.}
        \label{fig:recipe1}
    \end{minipage}
    \hspace{0.05\linewidth} 
    \begin{minipage}{0.47\linewidth}
        \centering
        \includegraphics[width=0.9\linewidth]{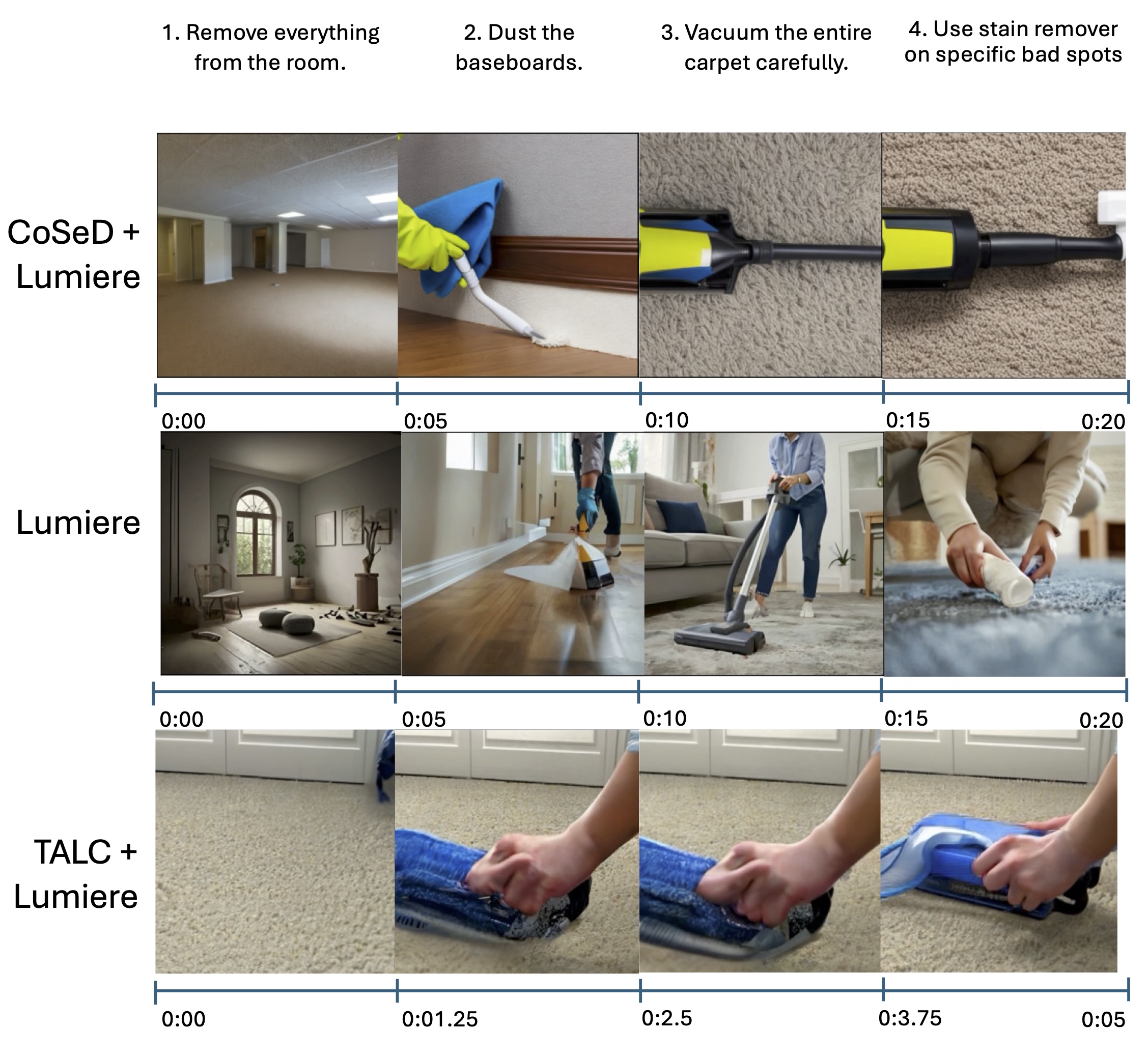}
        \caption{Example of an illustration for the DIY domain.}
        \label{fig:diy1}
    \end{minipage}

    \vspace{0.3cm}

    \centering
    \begin{minipage}{0.47\linewidth}
        \centering
        \includegraphics[width=0.9\linewidth]{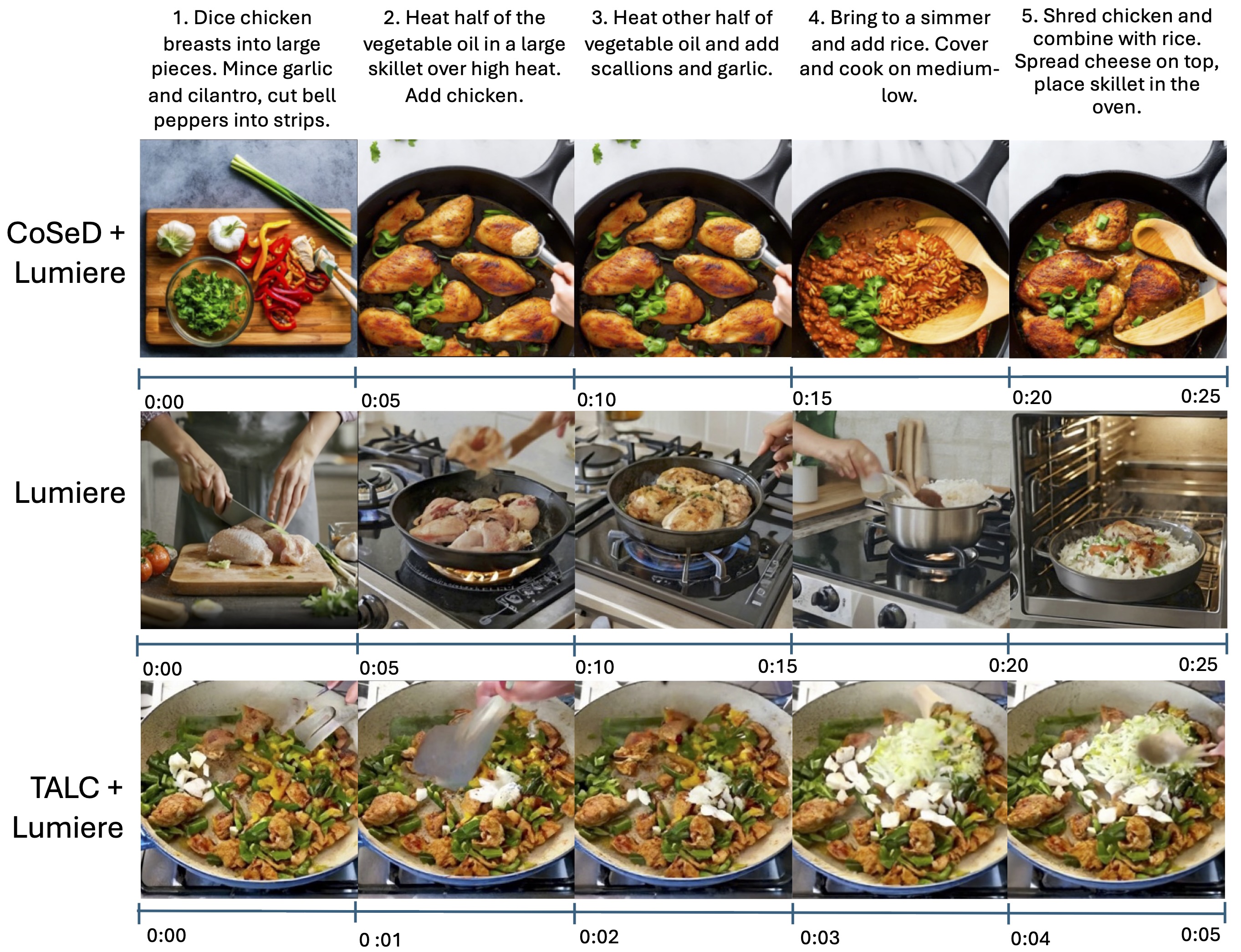}
        \caption{Example of an illustration for the recipe domain.}
        \label{fig:recipe2}
    \end{minipage}
    \hspace{0.05\linewidth} 
    \begin{minipage}{0.47\linewidth}
        \centering
        \includegraphics[width=0.9\linewidth]{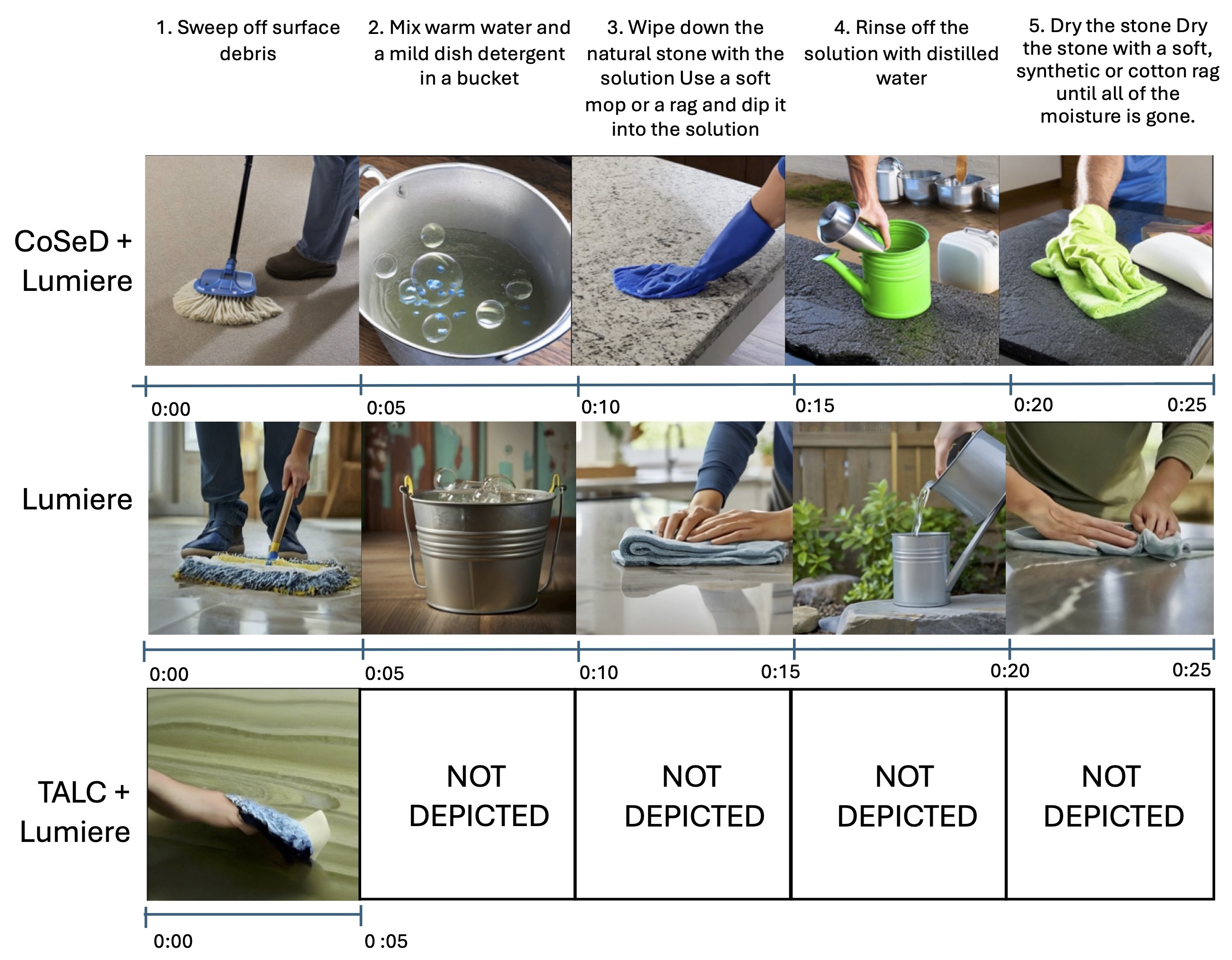}
        \caption{Example of an illustration for the DIY domain.}
        \label{fig:diy2}
    \end{minipage}
\end{figure*}

\section{Results and Discussion}

This section presents the evaluation results and discussion of our model's performance, with both human and automatic evaluations. We first examine the results of the human evaluation, followed by the automatic evaluation metrics. Finally, we discuss qualitative results and present ablation studies.

\subsection{Human evaluation}
The human evaluation was conducted with a focus on two primary criteria. Annotators were asked to assess 	\textbf{Semantic Consistency}, which measures how well the video matches the instructional text, and \textbf{Sequence Consistency}, which involves rating the video on text alignment, visual consistency and video quality. See the appendix file for details.

\paragraph{Multi-Scene Consistency Assessment.}

The results in Table~\ref{tab:manual-evaluation} show that \model combined with Lumiere achieves the highest Sequence Consistency at 74.2\% and leads in Semantic Consistency with 85.0\%, highlighting its effectiveness. This shows that the combination of \model and Lumiere is particularly effective for multi-scene generation tasks.

In contrast, methods such as TALC + ModelScope and TALC + Lumiere show significantly lower Semantic Consistency scores (38.3\% and 30.0\%, respectively) and a Sequence Consistency of 50.8\%. 
Although SD + SVD and Lumiere alone perform better, they still do not match the performance of \model + Lumiere, underscoring the advantages of our approach in achieving better coherence and text adherence in the generated videos.

\begin{table}[]
\centering
\small
\begin{tabular}{@{}lccc}
\toprule
\parbox[c]{1cm}{\centering \textbf{Methods}} & \parbox[c]{1cm}{\centering \textbf{Video\\Length}} & \parbox[c]{1.3cm}{\centering \textbf{Semantic\\Consist.}} & \parbox[c]{1.3cm}{\centering \textbf{Sequence\\Consist.}} \\ \midrule
\model + Lumiere & 20.8 s & \textbf{85.0} & \textbf{74.2} \\
\model + SVD & 14.9 s & 78.3 & 69.2 \\
TALC + ModelScope & 7.4 s & 38.3 & 50.8 \\
TALC + Lumiere & 5.0 s & 30.0 & 50.8 \\
SD + SVD & 14.9 s & 80.0 & 66.3 \\
Lumiere & 20.8 s & \underline{81.7} & \underline{72.9} \\ 
\bottomrule
\end{tabular}%
\caption{Manual evaluation of multi-scene video generation models based on two key criteria: \textbf{Semantic Consistency}, which measures the alignment of generated content with the described task steps, and \textbf{Sequence Consistency}, which assesses the visual coherence, text alignment, and overall quality of the video.}
\label{tab:manual-evaluation}
\end{table}

\paragraph{Videos Length.} We also report the length of the videos generated in Table~\ref{tab:manual-evaluation}. \model-based methods (\model + SVD and \model + Lumiere) achieve an average video length of around 15 and 21 seconds, respectively. This is substantially longer than TALC-based methods (TALC + ModelScope and TALC + Lumiere) which generate shorter videos, around 7 and 5 seconds on average. 

A visual inspection of the generated videos (Figure~\ref{fig:recipe1}) clearly indicates that \model successfully depicts all steps in the task, whereas TALC, despite being a multi-scene model, cannot achieve it. Even when TALC successfully depicts all steps, each scene typically lasts no more than 1.5 seconds, considering the average task length of 4.9 scenes (as detailed in the appendix file). In contrast, our model consistently achieves at least 3 seconds per scene, effectively providing double the duration for each step.

\paragraph{Side-by-Side Evaluation.}
To assess models prioritizing coherence against our best model, we conduct a side-by-side evaluation. Annotators choose which videos better represent task steps, directly comparing each model's coherence across scenes. See the appendix file for details.
This evaluation focuses on our best model, \model + Lumiere, against coherence-focused models such as TALC + ModelScope, TALC + Lumiere, \model + SVD, and the second-best model from Table \ref{tab:manual-evaluation}, Lumiere.

\begin{figure}[t]
     \centering
     \includegraphics[width=\linewidth]{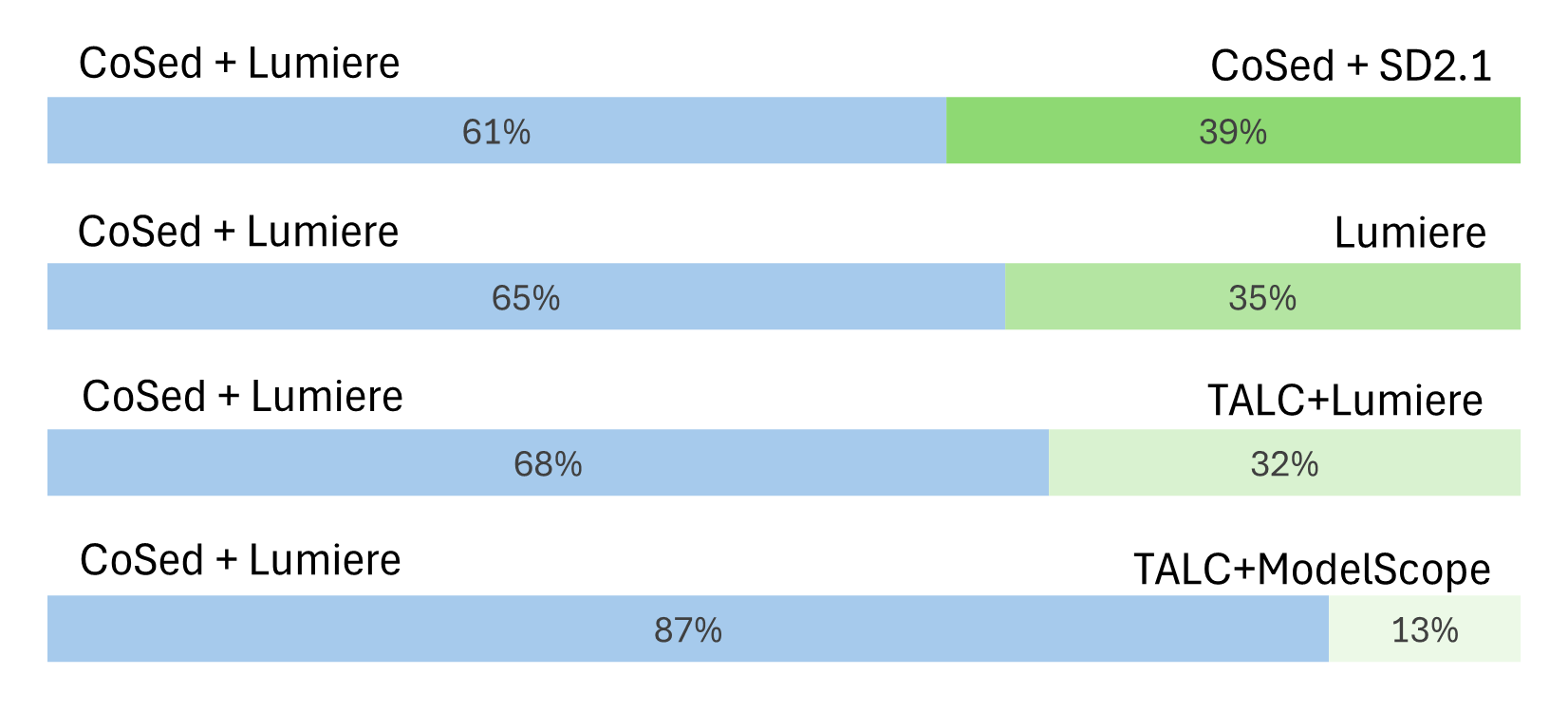}
     \caption{Annotators choice of coherent video generation models in a side-by-side comparison.}
     \label{fig:side-eval}
\end{figure}

According to the side-by-side evaluation results, Figure~\ref{fig:side-eval}, \model + Lumiere consistently outperforms all other models, with annotators repeatedly selecting it over competitors. For example, \model + Lumiere achieves a selection rate 87\% compared to just 13\% for TALC + ModelScope, demonstrating its superior ability to maintain coherence. It is also chosen 68\% of the time over TALC + Lumiere, which has a selection rate of 32\%, reflecting its better task consistency.

Compared to other \model variations, \model + Lumiere maintains its advantage. It is selected 61\% of the time over \model + SVD and outperforms Lumiere with a selection rate of 65\% versus 35\%. These results highlight its exceptional coherence in multi-scene tasks.

This evaluation clearly demonstrates that annotators consistently prefer \model + Lumiere over other models, highlighting its superior ability to maintain coherence across scenes and establishing it as the most effective model for managing multi-scene tasks.

\paragraph{\model vs Groundtruth.}

To evaluate the absolute quality of the generated video sequences, human annotators rated each sequence on a scale of 1 to 5, comparing them to ground-truth sequences. Deviations like hallucinated visual artifacts or inconsistent actions affect perceived quality. As shown in Table~\ref{tab:our-method-vs-ground-truth}, our method achieves more than 64\% of the ground truth score, with ground truth sequences scoring just 0.5 points below the maximum.

\begin{table}[h]
\centering
\small
\begin{tabular}{@{}lc@{}}
\toprule
\textbf{Method}       & \textbf{Average Rating}       \\ \midrule
\model+Lumiere   & 2.9 $\pm$ 0.99 \\
Ground-truth & 4.5 $\pm$ 0.55 \\ \bottomrule
\end{tabular}
\caption{Human annotation for the comparison of the proposed method with ground-truth scenes.}
\label{tab:our-method-vs-ground-truth}
\end{table}

\subsection{Automatic evaluation}

We employ CLIP~\cite{clip} to evaluate the sequence similarity of each task ($V \mapsto V$) and to assess the adherence of the generated image to the given textual descriptions ($T \mapsto V$). This provides a comprehensive evaluation of the alignment of the generated images with the intended textual descriptions and their visual coherence throughout the sequence.

\begin{table}[t]
    \centering
    \small

        \begin{tabular}{@{}clcc@{}}
            \toprule
            
& \textbf{Method} & $V \mapsto V$ & $T \mapsto V$ \\ \midrule
            \multirow{3}{*}{\rotatebox[origin=c]{90}{\footnotesize \textbf{Image}}} 
            & \model                   & 84.8                          & \textbf{27.1} \\
            & Seed-Llama               & 88.0                          & 16.5  \\
            & GILL                     & \textbf{88.2}                          & 22.3 \\ 
            \midrule
            \multirow{6}{*}{\rotatebox[origin=c]{90}{\footnotesize \textbf{Video}}} 
            & \model + SVD             & 84.8                         & \textbf{27.1}                \\
            & \model + Lumiere         & 84.8                          & \textbf{27.1}        \\
            & TALC + ModelScope        & 81.8                          & 12.4          \\
            & TALC + Lumiere           & \textbf{90.7}                 & 15.3          \\
            & SD + SVD                 & 82.1                          & 26.4 \\
            & Lumiere                  & 83.4                          & 14.9 \\ 
            \bottomrule
        \end{tabular}
    \caption{Automatic evaluation in terms of CLIP visual similarity ($V \mapsto V$) and CLIP semantic similarity ($T \mapsto V$).}
    \label{tab:automatic-evaluation}
\end{table}

\paragraph{\model Performance.}
\model achieves a sequence similarity score ($V \mapsto V$) of $84.8$ and a description adherence score ($T \mapsto V$) of approximately $27.1$ (see Table~\ref{tab:automatic-evaluation}), outperforming the image and video baselines in textual adherence. These results highlight \model's ability to generate sequences that are both visually coherent and semantically aligned with the text, demonstrating its significant potential for practical applications.

\paragraph{Comparison with Baseline Models.}
In a comparison to other sequence-generating models (see Table~\ref{tab:automatic-evaluation}), \model with both Video Stable Diffusion and Lumiere consistently achieves the highest description adherence ($T \mapsto V$) without discarding sequence similarity ($V \mapsto V$).

When comparing \model with the best model in sequence similarity, our model lags only $5.9\%$ while it gains $11.8\%$ in description adherence. Although TALC + Lumiere excels in maintaining high sequence similarity, \model demonstrates superior adherence to descriptions without compromising sequence similarity. This strong description adherence score highlights the effectiveness of our model in aligning generated content with text descriptions, which is crucial for tasks such as accurately converting textual descriptions into videos. Compared to vanilla video-only models, our approach surpasses both metrics, leading to improved results.

\subsection{Qualitative Analysis}
Figure~\ref{fig:recipe1} and~\ref{fig:recipe2} provide a closer look at how different methods influence the quality of generated videos and image sequences for a certain recipe. Our model excels in maintaining a consistent background, keeping the same pan, and ensuring that the ingredients evolve seamlessly from raw to final recipe. This clear depiction of the sequence provides viewers with a visually stable and easy-to-follow video.

\begin{figure*}[t]
    \centering
    \begin{minipage}{0.28\linewidth}
        \centering
        \includegraphics[height=70pt]{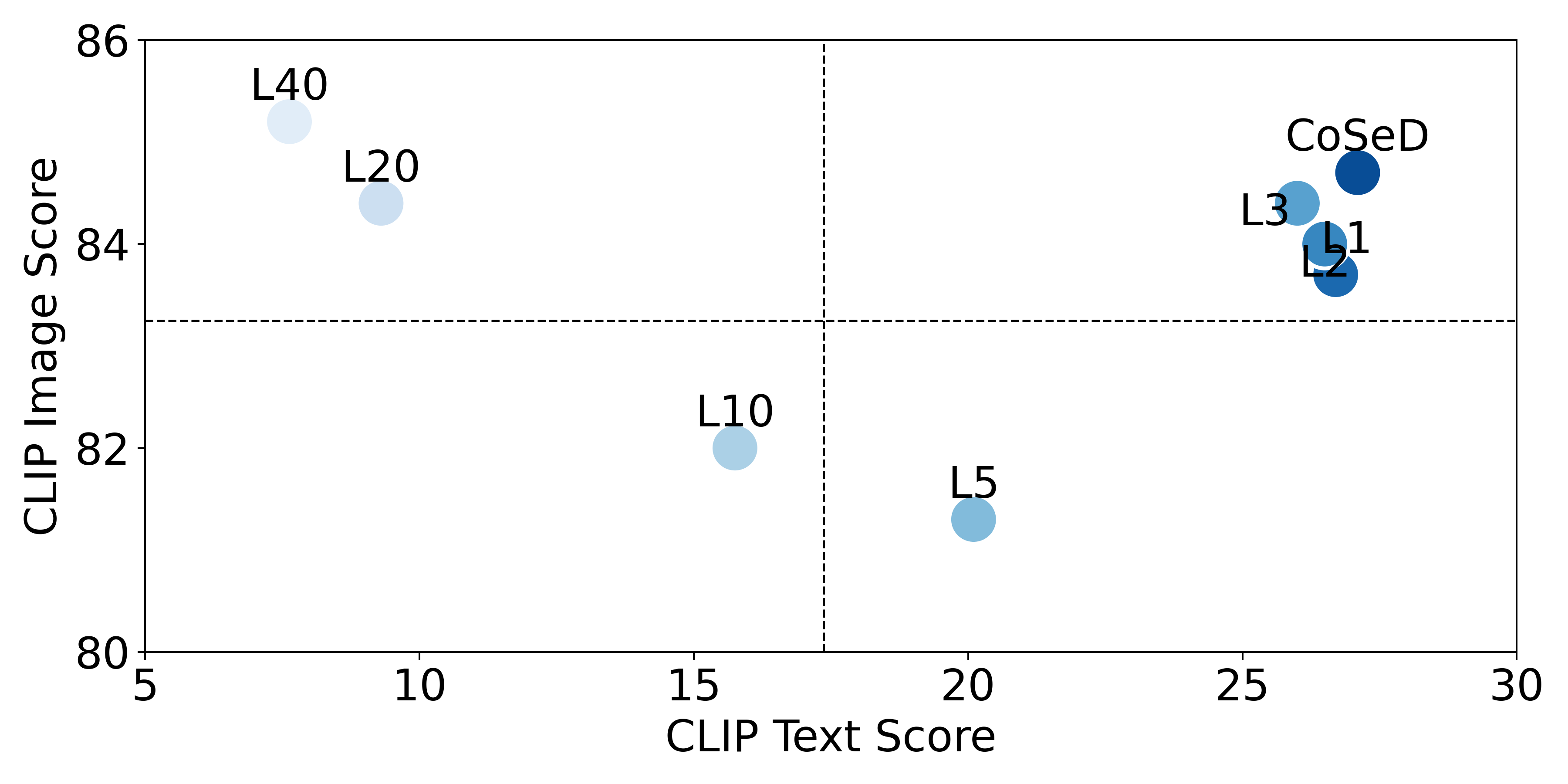}
        \caption{Impact of denoising latents in the performance of \model's visual and semantic similarity.}
        \label{fig:latent_evaluation}
    \end{minipage}
    \hspace{0.02\linewidth} 
    \begin{minipage}{0.36\linewidth}
        \centering
        \includegraphics[height=70pt]{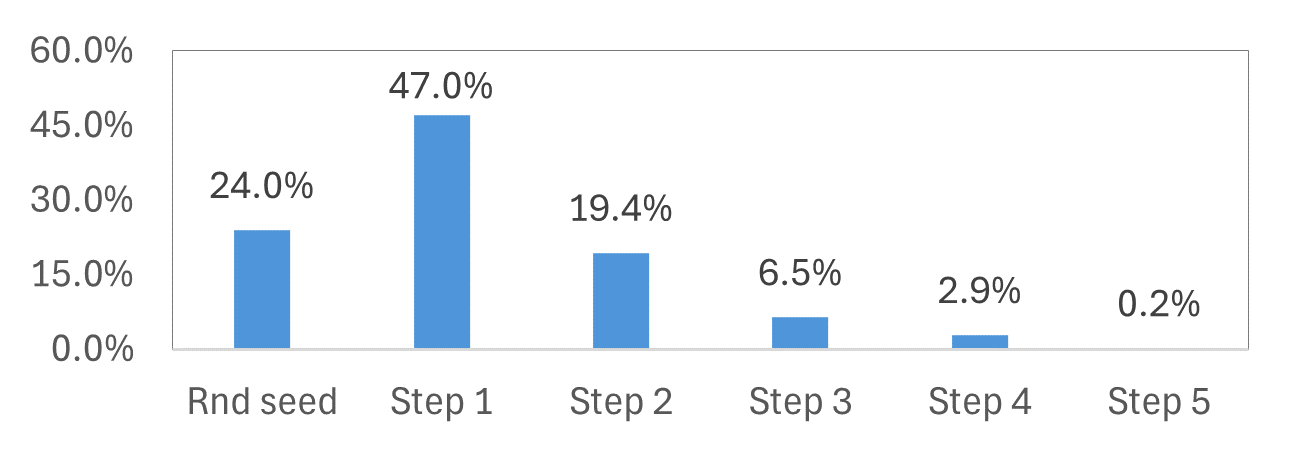}
        \caption{The average number of times that \model selected each task step for generating the next scene.}
        \label{fig:CoSed-step-usage}
    \end{minipage}
    \hspace{0.02\linewidth} 
    \begin{minipage}{0.3\linewidth}
        \centering
        \includegraphics[height=70pt]{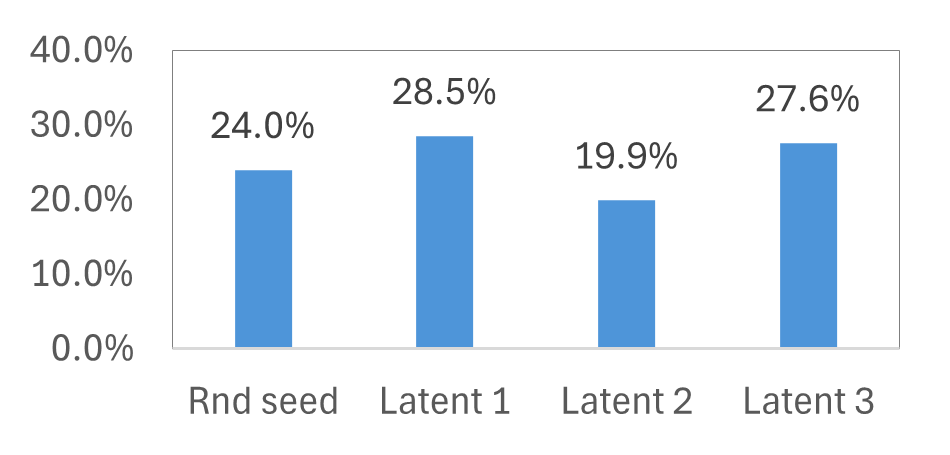}
        \caption{The average number of times that \model selected a given latent to generate the next scene.}
        \label{fig:CoSed-latent-usage}
    \end{minipage}
\end{figure*}

For the out-of-scope DIY tasks shown in Figures~\ref{fig:diy1} and~\ref{fig:diy2}, our model effectively handles broader actions, such as showing a room without furniture or a cleaning process, with good text adherence. However, it struggles with depicting complex tools such as vacuum cleaners. Other methods often generate very similar images, failing to accurately represent the task or omitting steps entirely, as demonstrated by TALC.

\subsection{Ablation Studies}
This section analysis \model contrastive selection of latents and steps along with its role in ensuring alignment with (possibly non-linear) sequences of instructions.

\paragraph{Denoising Latents.}
Understanding how latent variables affect sequence coherence and adherence is key to refining our model. Figure~\ref{fig:latent_evaluation} shows a correlation between latent settings and the model's ability to produce coherent, textually aligned sequences.
The analysis in Figure~\ref{fig:latent_evaluation} identifies four key latent configuration areas. The ideal zone balances high sequence similarity (CLIP Image Score) and description adherence (CLIP Text Score).
Latent 5 shows moderate text adherence, but lacks visual coherence. Latents 20 and 40 produce coherent images but deviate from task steps, leading to weak text adherence. Latent 10 performs poorly overall.

\begin{figure}[h]
    \centering
    \includegraphics[width=0.8\linewidth]{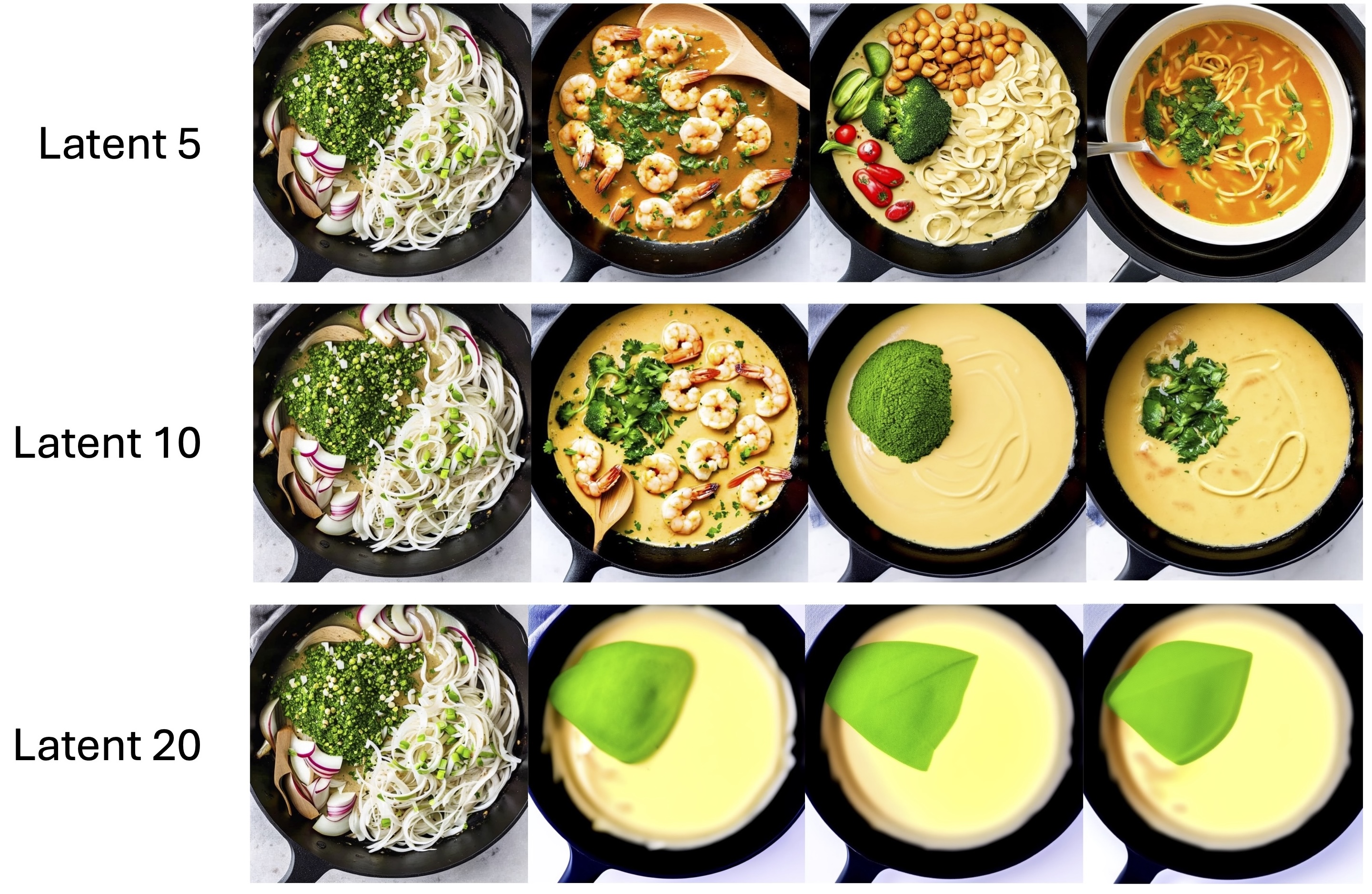}
    \caption{Impact of later latents on sequence generation coherence and text adherence.}
    \label{fig:latent_qualitative}
\end{figure}
Figure~\ref{fig:latent_qualitative} provides examples from these key latent areas: Latent 5 generates images that adhere to the prompt but lack coherence, Latent 10 starts with good coherence and text alignment but loses coherence over time, and Latent 20 produces overly similar images with low text adherence.

Ultimately, \model (top right mark in Figure~\ref{fig:latent_evaluation}) achieves superior performance by leveraging early denoising latents, using contrastive selection to enhance results compared to using individual latents.

\paragraph{Non-linear Video Scene Generation.}
A key feature of \model is its ability to evaluate denoising iterations across previous steps. As shown in Figure~\ref{fig:CoSed-step-usage}, \model exploits this non-linearity by selecting latents from various steps, rather than focusing solely on the immediately preceding one.

Similarly, Figure~\ref{fig:CoSed-latent-usage} indicates that the model does not always select the same latent. This variability suggests that different latents contribute with different information to the final generation, which is why \model chooses the most suitable latent rather than opting for the most recent one.

\section{Conclusion}

Generating multi-scene instructional videos for complex tasks, such as DIY projects and recipes, all while maintaining sequence coherence and accurate scene representation, is not a trivial feat.
The proposed method addresses these challenges with key contributions. First, we employ a decoder model to generate visual prompts from the sequence of instructions to ground the generation in a common context, thus ensuring better scene accuracy. Second, by conditioning the diffusion process on images from previous scenes, the method maintains coherence across scenes. 
Third, and more importantly, the \model's contrastive selection of the most consistent image enables the assessment of all previous steps. The result is the selection of the image that is most related to the overall sequence rather than just the preceding step. Additionally, the contrastive nature of \model enables the generation of non-linear sequences of video scenes, an exclusive feature of \model.

In addition, the compact design of the model helps to achieve efficient training and easy domain-specific fine-tuning, while its flexibility supports ensembles of diffusion models for optimal performance.

Evaluations confirm that our method effectively maintains scene coherence and accurately represents textual descriptions, as demonstrated by both manual and automatic evaluations. Importantly, side-by-side human annotation shows that annotators prefer our model over $65\%$ of the time, highlighting the effectiveness of our sequence-grounded approach. This preference underscores the value of our method in producing coherent and high-quality instructional videos.

\section*{Acknowledgements}
We thank anonymous reviewers for their valuable comments and suggestions. This work was partially supported by a Google Research Gift and by the FCT project NOVA LINCS Ref. (UIDB/04516/2020).

\clearpage

{\small
\bibliographystyle{ieee_fullname}
\bibliography{bibfile}
}

\clearpage
\appendix

\section{Dataset}
\label{ann:dataset_details}
Each task in the dataset includes a title, a description, a list of ingredients/resources and tools, and a sequence of step-by-step instructions, which may or may not be illustrated. To facilitate the illustration of task steps, we focused on tasks that are mostly illustrated, allowing us to use these images as the ground truth for training and evaluating our methods.

The dataset comprises approximately 1,400 tasks, with an average of 4.9 steps per task, which is a total of 6,860 individual steps. Most tasks include an image for each step, and some feature a complete recipe video that is segmented into multiple clips, with each clip lasting between 10 and 30 seconds per step.

Considering that the number of illustrations can affect the accuracy, we limited the training tasks to those with no more than 10 steps.

\section{Model Training}
\label{ann:appendix-training}

We opted for the CLIP model with a patch size of 32 to serve as the encoder for both image and text data due to its reputation in effectively capturing visual and textual information. In training our own architecture, we conducted experiments with various hyperparameters, including different learning rates, learning rate schedulers, dropout rates, layer freezing, and batch sizes, to identify the most suitable settings for our specific problem.

For the loss function, we employed cross-entropy, comparing the softmax output with hot-encoding of the steps that belong to each task. It is important to note that while this loss function indicates step-task associations, it may not always accurately reflect the model's overall performance on the task at hand.
During the tuning of hyperparameters, we found that freezing layers, weight decay, dropout, and learning rate schedulers had minimal impact on model performance.

\begin{table}[t]
\centering
\begin{tabular}{@{}ll@{}}
\toprule
\multicolumn{2}{c}{\textbf{Training Details}}         \\ \midrule
Optimizer         & Adam                              \\ 
Loss Function     & Cross-Entropy                     \\
Batch Size        & 500                               \\
Learning Rate     & 0.01                              \\
Epochs            & 10                                \\
Model Max Length  & 400                               \\ \midrule
Number of GPUs    & 1 A100-40GB                     \\ 
\bottomrule
\end{tabular}%
\caption{Training parameters}
\label{tab:training-specs}
\end{table}

The best model, which has about 600,000 parameters, was refined using specific training parameters listed in Table \ref{tab:training-specs}. Training was completed in under two minutes, using an A100-40GB GPU and spanning ten epochs. Employing the Cross-Entropy loss function, the training process operated with a batch size of 500 and a learning rate set at 0.01, using the Adam optimizer.

\paragraph{Single Modalities.}
In multimodal generation tasks, the integration of different modalities can notably impact the final output. Through this ablation study, we explore the implications of using singular modalities—text, images, or perturbed inputs—and examine the importance of modality mixing for enhancing generation quality.

Initially, in the scenario where text remains static across inputs, the model struggles with adaptability and generalization due to its reliance on a singular textual context. Conversely, when all inputs are randomized, the absence of consistent patterns across modalities impedes the model's learning process, resulting in suboptimal performance. However, the configuration where only text is randomized exhibits superior performance, suggesting that the model relies more on image over text. Notably, the marginal difference in performance between random text and the standard training approach underscores the intricate nature of multimodal tasks.

Our analysis underscores the importance of modality mixing in enhancing multimodal generation tasks. Integrating multiple modalities empowers the model to leverage diverse information sources, leading to more nuanced and accurate outputs.  In conclusion, the complexity of multimodal data show thats a model can relie more on one modality over the other but a mix of both will always be a better conjunction over a single modality.

\paragraph{Prompt Rewriter Training.}
The training process was centred on enabling the Large Language Model (LLM) to function as a visual caption generator for original task steps. Leveraging the capabilities of InstructBLIP~\cite{InstructBLIP}, we created contextual captions corresponding to each image and its associated step within the dataset. By integrating relevant task context into the generation of ground truth data, we enhanced the LLM's performance for visual clues. This approach ensured the production of accurate and contextually aligned visual descriptions, solidifying its role as an adept image caption generator.

\section{Human annotations}
\label{ann:human_eval}
The human annotation process conducted via Amazon Mechanical Turk evaluated multi-scene generated videos. Annotators rated visual quality, entity and background consistency, and text adherence using detailed guidelines.The process included instruction, qualification, and final evaluation phases, comparing our method's performance against other models and ground truth.

\begin{figure*}[t]
    \centering
    \includegraphics[width=\linewidth]{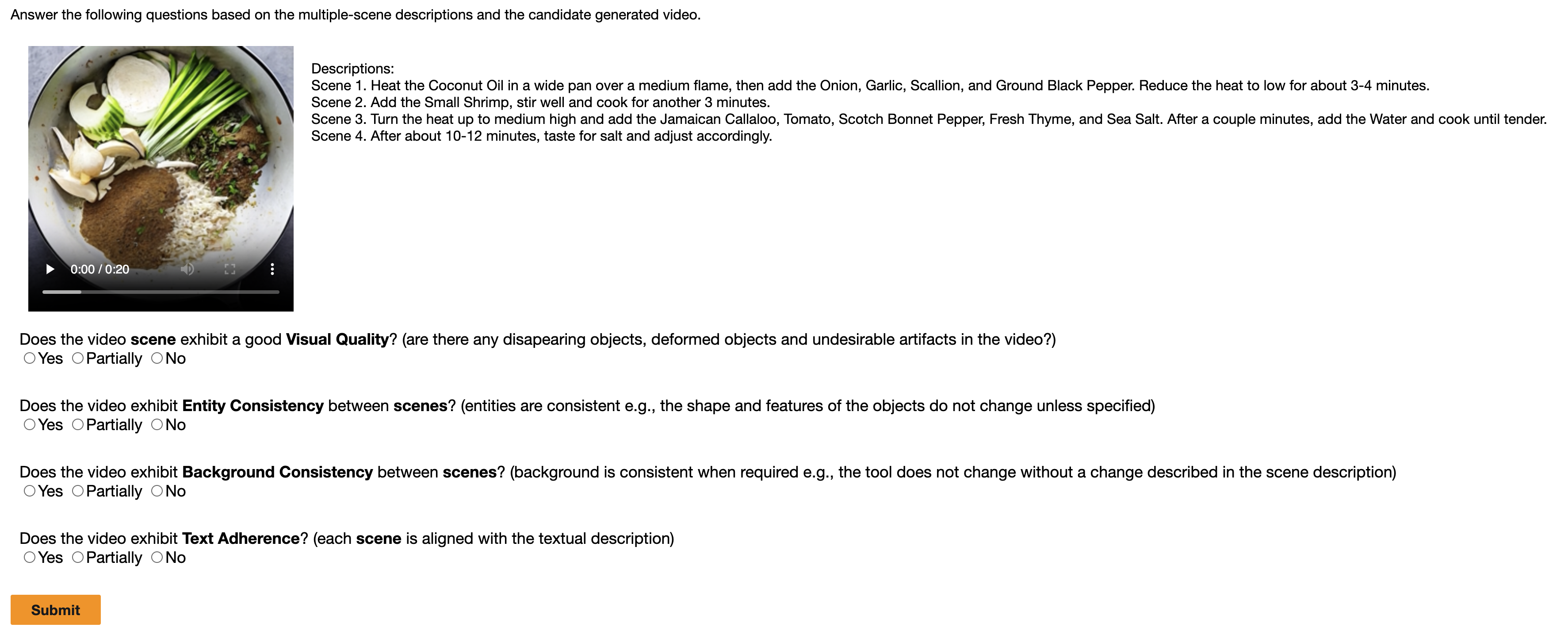}
    \caption{Human Annotation Layout for Video Generation Methods}
    \label{fig:annotation}
    \vspace{0.5em}
\end{figure*}

\begin{figure*}[t]
    \centering
    \includegraphics[width=\linewidth]{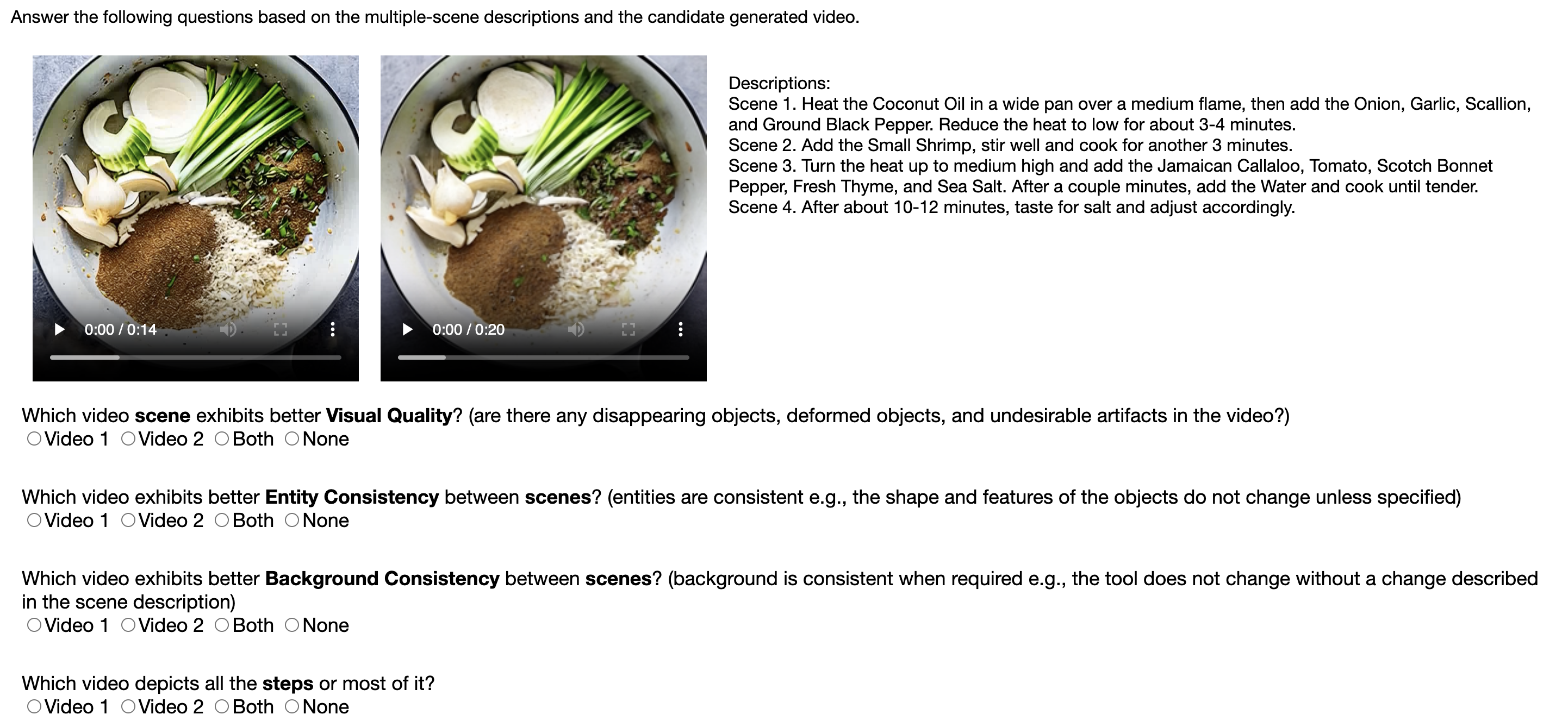}
    \caption{Side by Side  Annotation Layout for Video Generation Methods}
    \label{fig:annotation_side_by_side}
    \vspace{0.5em}
\end{figure*}

\begin{figure*}[t]
    \centering
    \fbox{
        \begin{minipage}{0.9\textwidth}
            \textbf{Instructions}

            \begin{enumerate}
                \item Watch the entire video provided on the left side of the screen.
                \item Carefully read the descriptions provided on the right side of the screen.
                \item Evaluate the video based on the following criteria:
                \begin{itemize}
                    \item \textbf{Visual Quality:} Check if the video scene has good visual quality without any disappearing or deformed objects and no undesirable artifacts.
                    \item \textbf{Entity Consistency:} Ensure that the entities (objects) are consistent between scenes, with no unexpected changes unless specified in the descriptions.
                    \item \textbf{Background Consistency:} Confirm that the background remains consistent between scenes, unless a change is described in the scene description.
                    \item \textbf{Text Adherence:} Verify that each scene in the video aligns with the corresponding textual description.
                \end{itemize}
                \item Select the appropriate answer for each question below the video and descriptions.
                \item Double-check your answers before submitting the form.
            \end{enumerate}
        \end{minipage}
    }
    \caption{Instructions for Video Generation Evaluation}
    \label{fig:annotation_guidelines_methods}
\end{figure*}

\subsection{Annotations Job}

Participants for this evaluation were recruited through the crowdsourcing platform Amazon Mechanical Turk. Annotators were compensated at a rate of \$0.5 per task, and each task was designed to take between 2 and 3 minutes to complete.

The payment rate of \$0.5 per task was determined based on pilot tests to estimate the average time required for completion and to ensure fair compensation for participants' time and effort. At this rate, annotators could earn approximately \$10 per hour if tasks were completed consistently within 3 minutes each, which exceeds the current federal minimum wage in the United States.

All annotators were aware that they were collaborating with researchers for an evaluation on video generation. They received detailed information about their tasks and how their evaluations would contribute to the research.

With focus on the task itself, we maintained annotators' anonymity. Consequently, we do not have specific demographic or geographic information about the annotators.

\subsection{Annotation Process}

The annotation process consisted of three main steps:

\begin{enumerate}
    \item \textbf{Instruction Phase}: Annotators received a slideshow with detailed instructions on how to perform the annotations. This phase included several examples to train the annotators and ensure clarity regarding the task requirements.
    \item \textbf{Qualification Phase}: After the instruction phase, annotators completed a qualification task involving five example annotations. This step was designed to assess their understanding and ability to perform the tasks according to our standards. Only those who passed this qualification phase proceeded to the final annotation phase.
    \item \textbf{Annotation Phase}: Qualified annotators were then given the full set of annotation tasks, where they evaluated the final results.
\end{enumerate}

\subsection{Annotation Tasks}

The human annotation pool consisted of annotators who successfully passed the qualification phase. 

Figure~\ref{fig:annotation} illustrates the task layout for selecting the best visual coherence maintaining method. Annotators evaluated six models: \textit{\model + Stable Video Diffusion}, \textit{\model + Lumiere}, \textit{TALC + ModelScope}, \textit{TALC + Lumiere}, \textit{Lumiere}, and \textit{Stable Diffusion + Stable Video Diffusion}. They graded the videos on visual quality, entity consistency, background consistency, and adherence to text. The same annotation method was used to compare our best method with other methods as seen in Figure~\ref{fig:annotation_side_by_side}.

Figure~\ref{fig:annotation_guidelines_methods} outlines the annotation guidelines for evaluating the sequences generated by our method compared to other baselines. This figure provides detailed criteria for the annotators to follow, ensuring consistency in their assessments.

\begin{figure*}[t]
    \centering
    \includegraphics[width=\linewidth]{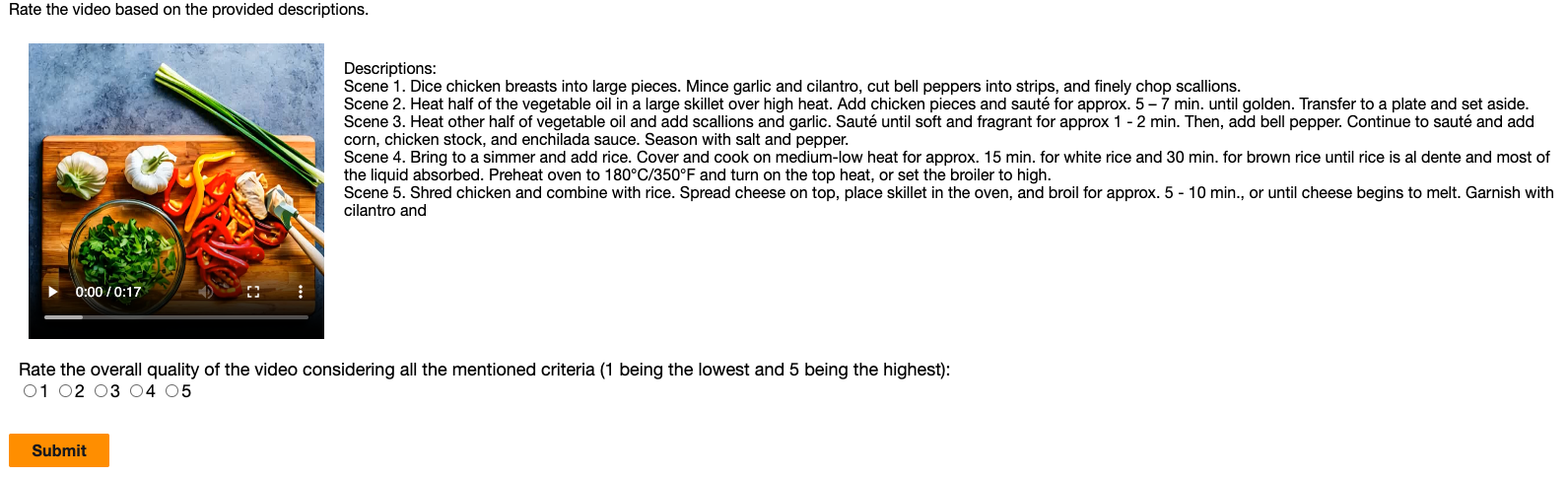}
    \caption{Human Annotation Layout for Our Method vs. Ground Truth}
    \label{fig:our-method-vs-ground-truth-annotations}
    \vspace{0.5em}
\end{figure*}

In Figure~\ref{fig:our-method-vs-ground-truth-annotations}, the specific task of rating sequences generated by our method against ground-truth images is depicted. Annotators were asked to score these sequences on a scale from 1 to 5, providing a quantitative measure of our model's performance.

\begin{figure*}[htbp]
    \centering
    \fbox{
        \begin{minipage}{0.9\textwidth}
            \textbf{Instructions}

            We will present you with a video clip representing a sequence of steps.

            Your task is to rate the video on a  scale of 1-5 based on the following factors:

            \begin{itemize}
                \item \textbf{Representation of Instructions:} How well does the video illustrate the given instructions?
                \begin{itemize}
                    \item \textit{Note:} Any generation artifacts should not impact the rating if the overall video clearly conveys the steps.
                \end{itemize}
                \item \textbf{Coherence:} How coherent is the sequence of scenes in the video?
                \begin{itemize}
                    \item \textit{Example:} If an object is blue in one scene, it should remain blue in subsequent scenes.
                    \item \textit{Example:} The background should remain consistent across the video.
                \end{itemize}
            \end{itemize}

        \end{minipage}
    }
    \caption{Instructions for Ground Truth Annotation}
    \label{fig:our-method-vs-ground-truth-annotation-guidelines}
\end{figure*}

To complement this, Figure~\ref{fig:our-method-vs-ground-truth-annotation-guidelines} presents the detailed guidelines used for rating sequences generated by our method compared to ground-truth images. These guidelines helped standardize the evaluation process, ensuring that the ratings were fair and consistent across different annotators.

\section{Prompt Optimization}
\label{ann:prompt}
We attempted to enhance generation quality by refining our prompts, incorporating detailed descriptions. These descriptions included: 

\begin{itemize}
\item \textbf{Main Subject}: Highlighting the primary focus of the image, whether it's ingredients in a recipe or materials for a project.
\item \textbf{Item}: Describing all inanimate objects, ranging from everyday items like utensils or tools to more abstract entities like machinery.
\item \textbf{Setting}: Depicting the broader environment or backdrop, spanning from kitchen countertops to workshop benches or outdoor landscapes.
\item \textbf{Activity}: Illustrating dynamic actions or steps that animate the imagery, such as stirring ingredients or assembling components.
\item \textbf{Arrangement}: Describing the spatial layout, indicating how elements are positioned relative to each other, like 'stacked neatly' or 'arranged in a circular pattern.'
\end{itemize}

In the end, though, these prompts failed to produce better outcomes.

\section{Selecting the First Image}
\label{ann:human_selection}

In the process of generating visual representations based on textual input, the selection of the initial image or video is crucial. This selection not only serves as the first interaction with the user but also influences subsequent representations, directly impacting the overall quality of the generated content. Therefore, establishing a robust strategy for selecting the first image is essential to ensure coherence and effectiveness in the generated output.

The significance of the initial image choice lies in its potential to enhance user engagement. A mismatch between the text and visual representation can disrupt comprehension and decrease the overall user experience. Therefore, the selection strategy should consider factors such as alignment with the text, diversity, and relevance to ensure a seamless transition from text to visuals.

\paragraph{Single Image Generation.}
This strategy offers simplicity and directness as its main advantages. By generating a single image, it provides a straightforward solution without added complexity. However, it may suffer from a lack of variety, potentially resulting in limited diversity in the initial representation. Additionally, its reliance on the Stable Diffusion model's capabilities means that the quality and text adherence of the generated image depends solely on the model's performance.

\paragraph{Random Selection from Image Batch.}
The random selection strategy offers increased diversity and reduced bias. By allowing the selection of a random image from a batch, it potentially offers a wider range of visual representations. Moreover, it avoids intentional or unintentional bias in selecting the first image. However, it lacks control over the selection process, which may lead to the choice of an image that does not align well with the text. Furthermore, the quality and relevance of the selected image may vary across different runs, introducing potential inconsistency.

\paragraph{Using CLIP for Selection.}
In this approach, CLIP is used to meticulously select the initial image based on its semantic similarity to the textual description. By leveraging CLIP's robust understanding of semantics, we ensure that the chosen image corresponds well with the text, thereby enhancing both coherence and relevance. Importantly, the computational overhead associated with CLIP is minimal compared to the resource-intensive task of image generation. Unlike other options, utilizing CLIP for image selection notably increases the likelihood of achieving coherence and relevance in the first generated content.

\begin{figure}[h]
    \includegraphics[width=\linewidth]{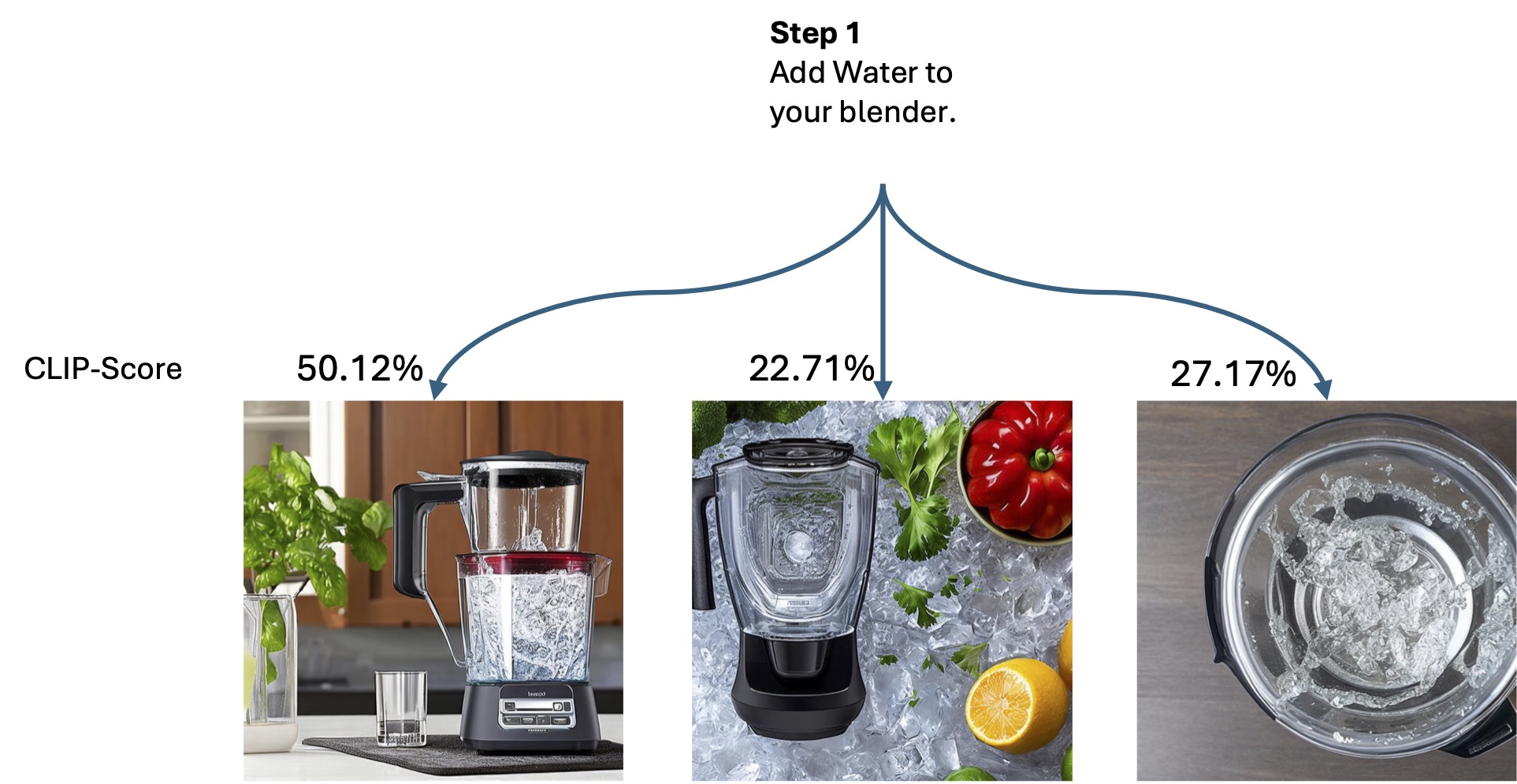}
    \caption{CLIP Selection}
    \label{fig:clip_selection}
\end{figure}

Selecting the first image during the generation of visual representations from sequential text input is a critical step. Among the presented strategies for selecting the initial image, the approach of using CLIP for selection stands out as the most promising. This strategy emphasizes semantic alignment, ensuring coherence and relevance between the text and visual representation. By leveraging CLIP's capabilities, we aim to enhance the overall quality and effectiveness of the generated output.

\end{document}